\documentclass[11pt]{article}

\usepackage[final]{acl}

\usepackage{times}
\usepackage{latexsym}

\usepackage[T1]{fontenc}

\usepackage[utf8]{inputenc}

\usepackage{microtype}
\usepackage{amsmath}
\usepackage{algorithm}
\usepackage{algorithmic}
\usepackage{inconsolata}
\usepackage{cleveref}
\usepackage{graphicx}
\usepackage{booktabs}
\usepackage{xcolor}
\usepackage{amssymb}

\newcommand{\token}[1]{\textbf{\texttt{#1}}}

\title{Calibrated Speculative Decoding: Frequency-Guided Candidate Selection for Efficient Inference}

\author{
  \textbf{Xuwen Zhou}\textsuperscript{1}, \textbf{Fangxin Liu}\textsuperscript{1*}, \textbf{Chao Wang}\textsuperscript{2}, \textbf{Xiao Zheng}\textsuperscript{2},\\
  \textbf{Hao Zheng}\textsuperscript{2}, \textbf{Min He}\textsuperscript{2}, 
  \textbf{Li Jiang}\textsuperscript{1*}, \textbf{Haibing Guan}\textsuperscript{1} \\
  \textsuperscript{1}Shanghai Jiao Tong University / \textsuperscript{2}Alibaba Cloud Computing \\
  \texttt{\{xvenzhou, liufangxin, ljiang\_cs
  , hbguan\}@sjtu.edu.cn} \\
  {\textsuperscript{*}Corresponding authors}
}

\begin{document}
\maketitle
\begin{abstract}
Speculative decoding accelerates autoregressive generation by letting draft tokens bypass full verification, but conventional frameworks suffer from frequent false rejections, particularly when draft models produce semantically correct but lexically divergent outputs. In this paper, we present \textbf{Calibrated Speculative Decoding (CSD)}, a training-free framework that recovers valid tokens discarded by standard verification. Guided by the principle of \emph{``Frequency-Guided Candidate Selection, Probability-Guarded Acceptance,''} CSD incorporates two lightweight modules: \textbf{Online Correction Memory}, which aggregates historical rejections to propose recurring divergence patterns as rescue candidates, and \textbf{Semantic Consistency Gating}, which verifies candidate admissibility using probability ratios instead of exact token matching. Our evaluation across diverse large language models demonstrates that CSD outperforms existing methods, achieving a peak throughput speedup of $2.33\times$. CSD preserves model accuracy across all tasks while boosting performance on complex reasoning datasets. These results establish CSD as a highly effective, lightweight solution for LLM deployments.
\end{abstract}

\section{Introduction}
\label{sec:intro}

\begin{figure*}[t] 
\vspace{-0.5cm}
\setlength{\abovecaptionskip}{3pt}
\setlength{\belowcaptionskip}{3pt}
  \centering
  \includegraphics[width=0.95\textwidth]{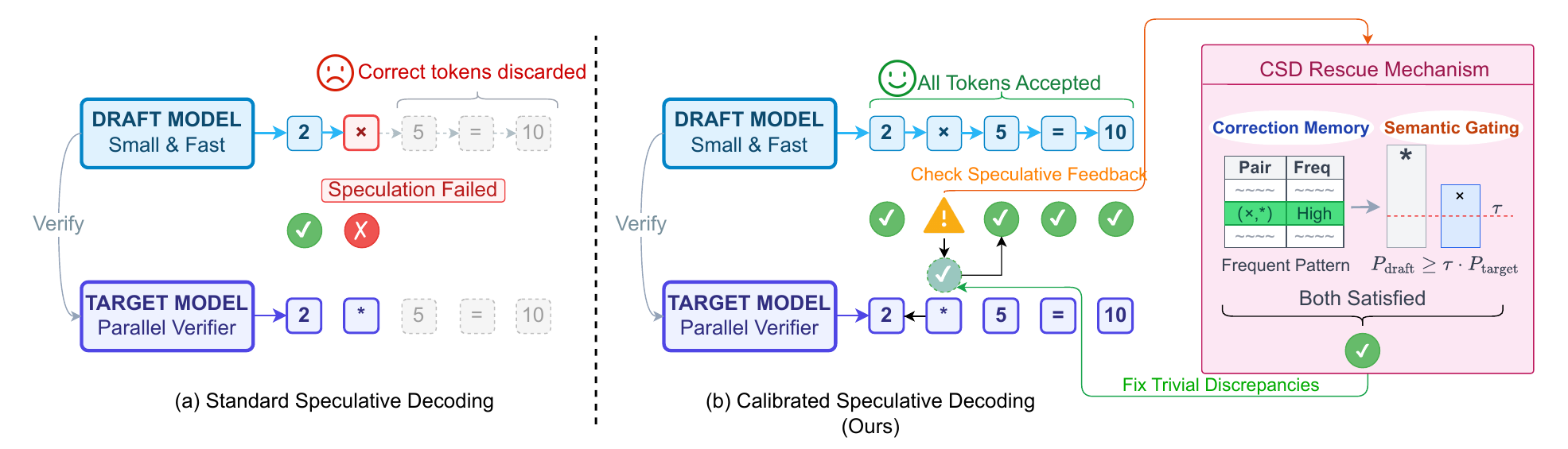} 
\caption{Overview of Calibrated Speculative Decoding (CSD). 
\textbf{(a)} Standard Speculative Decoding suffers from ``False Rejections,'' where trivial lexical mismatches (e.g., `x' vs `*') cause the discard of subsequent valid tokens (e.g., \texttt{5=10}). 
\textbf{(b)} CSD mitigates such rigid rejections by introducing a Rescue Mechanism. It employs Online Correction Memory (OCM) to propose potential candidates based on historical priors, and uses Semantic Consistency Gating (SCG) to verify their admissibility via probability ratios (Score $\ge \tau$), effectively recovering wasted computation.}
\label{fig:overview}
\end{figure*}

Large Language Models (LLMs) have demonstrated unprecedented capabilities across a wide spectrum of tasks \citep{achiam2023gpt, guo2025deepseek}. However, their deployment is severely constrained by the memory wall \citep{gholami2024ai}. Specifically, the autoregressive nature of decoding renders inference predominantly memory-bandwidth bound, particularly in latency-sensitive, small-batch settings, where the arithmetic intensity is low \citep{chen2025towards, pope2023efficiently,liu2024spark,liu2021improving}.

To alleviate this bottleneck, \textbf{Speculative Decoding (SD)} \citep{leviathan2023, chen2023accelerating} has emerged as a dominant acceleration paradigm. SD employs a smaller draft model to propose candidate tokens that are verified in parallel by the target model, thereby amortizing memory access costs across multiple tokens and enabling lossless acceleration.

Recent advances in Small Language Models (SLMs) alter the role of draft models in SD. 
Driven by advances in model distillation and architectural optimization, modern SLMs are no longer mere approximators but capable reasoners \citep{bachmann2025judge}. For instance, Llama-3-8B achieves 84.5\% accuracy on the GSM8K math benchmark, significantly outperforming its much larger predecessor, Llama-2-70B (56.8\%) \citep{dubey2024llama, touvron2023llama}. This paradigm shift presents a new opportunity: draft models are increasingly generating semantically correct responses that may differ lexically from the target model's preference.

However, existing SD frameworks fail to capitalize on this intelligence. Standard verification mechanisms, such as strict token matching or rejection sampling \citep{leviathan2023}, operate on a rigid, distribution-alignment basis. This creates a fundamental mismatch: while the draft model demonstrates strong reasoning capabilities, the verification logic remains structurally intolerant. As illustrated in \Cref{fig:overview}(a), this rigidity leads to ``False Rejections,'' where valid draft tokens (e.g., synonyms like \token{`x'} vs \token{`*'}) are discarded merely due to trivial lexical or stylistic discrepancies. This forces the target model to re-generate semantically equivalent tokens, thereby negating the efficiency gains, particularly in open-ended generation tasks.

To bridge the gap between rigid verification and flexible semantics, we introduce \textbf{Calibrated Speculative Decoding (CSD)}, a training-free framework designed to recover valid tokens from false rejections. Guided by the philosophy of \emph{``Frequency-Guided Candidate Selection, Probability-Guarded Acceptance,''} CSD incorporates two lightweight modules (\Cref{fig:overview}(b)). First, Online Correction Memory continuously aggregates historical rejections to propose ``rescue candidates'' for recurring divergence patterns. Second, Semantic Consistency Gating evaluates the admissibility of these candidates through probability ratios rather than exact matching.

Our contributions are as follows: (1) We identify the Semantics-Alignment Mismatch in modern speculative decoding, revealing that rigid verification hinders the potential of draft models; 
(2) We propose Calibrated Speculative Decoding (CSD), a training-free framework that incorporates Online Correction Memory and Semantic Consistency Gating to enable adaptive, pattern-aware verification; 
(3) CSD delivers up to $2.33\times$ speedup while maintaining generation quality across standard tasks. Importantly, it uniquely enhances complex reasoning accuracy by 2.5 points on HumanEval and 2.0 points on MATH500~\cite{hendrycks2021measuring, chen2021evaluating}, proving that CSD safely recovers valid draft tokens to improve inference efficiency.

\section{RELATED WORK}
\label{sec:rw}

Speculative Decoding (SD) breaks the serial dependency of autoregressive generation by introducing a drafting step, establishing itself as a dominant paradigm for accelerating LLM inference. The classic approach typically employs an independent lightweight model from the same family as the target model to generate draft tokens \citep{chen2023accelerating, leviathan2023}. To circumvent the VRAM overhead and system complexity associated with independent models, some works utilize layer skipping or early-exiting strategies to construct draft models \citep{elhoushi2024layerskip, zhang2024draft,xia2024swift,liu2024speculative}, reusing a subset of target weights for efficient generation. Meanwhile, approaches like Eagle \citep{li2024eagle} and Medusa \citep{cai2024medusa} append lightweight decoding heads to the target model to predict future tokens in parallel. Furthermore, methods such as Lookahead Decoding \citep{fu2024break} and Jacobi decoding \citep{santilli2023accelerating} eliminate the draft model entirely, achieving parallelism via parallel iterative decoding.

To maximize the speedup of SD, prior works have focused on improving the acceptance rate. 
On one hand, Tree-based methods \citep{miao2024specinfer,chen2024sequoia} extend single-chain generation to token trees built from top-$k$ predictions and verify them in parallel with tree-attention, enlarging the covered solution space.
On the other hand, efforts improve draft quality through alignment strategies, such as distilling the draft from the target model \citep{zhou2023distillspec} or sharing KV caches across models \citep{du2024glide}. Recently, approaches such as Judge Decoding \citep{bachmann2025judge} and R2R \citep{fu2025r2r} explore learned verification by training auxiliary models to assess token validity.

However, these advancements often introduce substantial overheads, including the high computational cost of tree-based verification and the training expense of learned methods, while remaining constrained by rigid exact matching. Such rigidity leads to the false rejection of semantically equivalent tokens.
In contrast, we propose a lightweight, training-free framework that calibrates the verification logic itself. By leveraging historical divergence patterns and semantic gating, our approach recovers valid tokens without architectural modifications. Notably, our framework is generically applicable to standard rejection sampling-based methods, serving as a seamless enhancement to their verification logic.

\section{Preliminaries and Motivation}
\label{sec:Pre}

\subsection{Speculative Decoding (SD)}
Speculative Decoding accelerates the inference of a target model $\mathcal{M}_p$ by leveraging a computationally efficient draft model $\mathcal{M}_q$. The process follows a predict-then-verify paradigm. Given an input context $x$, the draft model first generates a sequence of $\gamma$ candidate tokens $\tilde{x}_{1:\gamma}$ autoregressively. Specifically, at each step $i$, the token $\tilde{x}_i$ is sampled via:
\begin{equation}
    \tilde{x}_i \sim q(\cdot \mid x, \tilde{x}_{<i}), \quad i \in \{1, \dots, \gamma\}
\end{equation}
where $q(\cdot)$ denotes the probability distribution output by $\mathcal{M}_q$. 

Subsequently, $\mathcal{M}_p$ evaluates the drafted sequence in a single parallel forward pass to compute the target probabilities $p(\cdot \mid x, \tilde{x}_{<i})$ for all positions. To ensure the generated sequence follows the target distribution, SD adopts a rejection sampling scheme. A candidate token $\tilde{x}_i$ is accepted with probability $\alpha_i$, defined as:
\begin{equation}
\label{eq:sd_acceptance}
    \alpha_i = \min\left(1, \frac{p(\tilde{x}_i \mid x, \tilde{x}_{<i})}{q(\tilde{x}_i \mid x, \tilde{x}_{<i})}\right)
\end{equation}
If a token $\tilde{x}_i$ is rejected, the process terminates at step $i$, and a correction token is resampled from the normalized residual distribution $p' = \text{norm}(\max(0, p - q))$. Conversely, if all $\gamma$ tokens are accepted, an additional token is sampled directly from $\mathcal{M}_p$. Crucially, this mechanism guarantees that the final output distribution is mathematically identical to that of the target model $\mathcal{M}_p$.

\begin{figure*}[t]
\vspace{-0.5cm}
\setlength{\abovecaptionskip}{3pt}
\setlength{\belowcaptionskip}{3pt}
  \centering
  \includegraphics[width=0.95\linewidth]{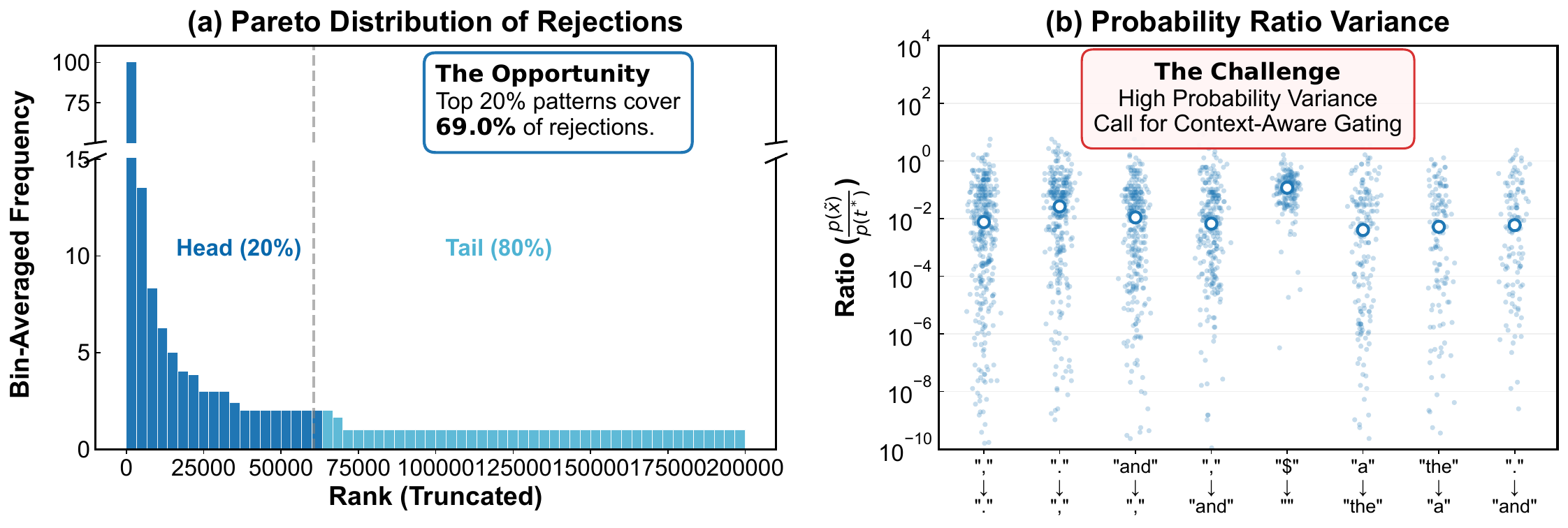} 
\caption{Statistical Analysis of False Rejections. 
    \textbf{(a)} Long-Tailed Frequency: The distribution of rejected patterns is highly skewed; the 20\% (head patterns) account for 69.0\% of total rejections. This suggests a lightweight Correction Memory targeting these patterns can recover most failed speculations.
    \textbf{(b)} Ratio Variance: 
    Extreme variance in probability ratios ($T=1$) reflects a spectrum from severe hallucinations to semantically consistent alternatives. This diversity necessitates dynamic Semantic Consistency Gating to distinguish valid candidates from errors, moving beyond static relaxation.}
    \label{fig:motivation}
\end{figure*}

\subsection{Why Do We Reject? Analyzing Draft Rejection Patterns}
\label{sec:empirical_analysis}
Rejection sampling guarantees correctness under exact token-level verification (Eq.~\ref{eq:sd_acceptance}), but this strict matching rule does not account for the lexical diversity of natural language. Recent studies~\citep{bachmann2025judge, fu2025r2r} show that draft and target models often differ by semantically equivalent lexical choices, referred to as ``neutral differences.'' Existing approaches typically relax rigid verification by introducing learned routers or judgment heads, which increases system complexity and inference overhead. In this work, we ask whether a subset of these valid but rejected draft tokens can be recovered by refining the verification logic in a training-free manner.

To characterize the prevalence and structure of such false rejections, we conduct a large-scale empirical analysis under standard speculative decoding. We use \texttt{LLaMA-3.2-1B-Instruct} as the drafter and \texttt{LLaMA-3-70B-Instruct} as the target, and collect all rejected token pairs during standard sampling ($T=1$). Experiments are performed on a combined dataset of 28,000 samples from CNN/DailyMail~\citep{hermann2015teaching} and MATH~\citep{hendrycks2021measuring}. \Cref{fig:motivation} summarizes the resulting statistics. We highlight two observations that inform our design.

\paragraph{Observation 1: Long-tailed distribution of rejection patterns.}
As shown in ~\Cref{fig:motivation}(a), rejected draft--target token pairs exhibit a highly skewed distribution. The top 20\% most frequent patterns account for approximately 69.0\% of all rejections. This concentration indicates that a limited number of recurring patterns dominate draft failures, suggesting that many rejections arise from systematic rather than isolated mismatches.

\paragraph{Observation 2: Strong context dependence within frequent patterns.}
Frequency alone, however, is insufficient to determine whether a rejected draft token should be accepted. \Cref{fig:motivation}(b) analyzes the top-10 most frequent patterns and reports distribution of probability ratios assigned by the target model to the rejected draft token $\tilde{x}_{i}$ versus the optimal token $t^*$ ($p_{\mathcal{M}_p}(\tilde{x}_{i}) / p_{\mathcal{M}_p}(t^*)$). For the same token pair, this ratio spans several orders of magnitude, ranging from values close to $1$ to below $10^{-10}$. This wide variance reflects strong context dependence. In some contexts, the draft token remains a plausible semantic alternative with high target confidence (e.g., \texttt{a} vs. \texttt{the}), while in others it corresponds to a factual error or hallucination that the target model appropriately suppresses.

Taken together, these observations show that exact matching is overly restrictive, discarding high-confidence draft tokens that are semantically valid, while purely frequency-based relaxation is unsafe due to context-sensitive failure cases. This motivates a verification strategy that leverages historical recurrence while enforcing confidence-aware gating, enabling selective recovery of valid draft tokens without introducing learned predictors or compromising correctness.

\section{Methodology}
\label{sec:Method}

\subsection{CSD Overview}
Building upon the insights from Section~\ref{sec:empirical_analysis}, we introduce \textbf{Calibrated Speculative Decoding (CSD)}, a training-free inference framework that improves the efficiency of speculative decoding by selectively relaxing rigid token-level verification.

As illustrated in \Cref{fig:overview}(a), 
conventional speculative decoding enforces exact token matching between the draft and target models.  This strict verification can lead to unnecessary rejections when the two models produce lexically different yet functionally comparable tokens in context (e.g., discarding \token{5=10} due to a trivial \token{x} vs.\ \token{*} mismatch), 
thereby reducing the effective acceptance rate and wasting computation.

To address this limitation, CSD adopts a dual-stage inference protocol guided by the principle of \emph{``Frequency-Guided Candidate Selection, Probability-Guarded Acceptance.''} As shown in \Cref{fig:overview}(b), this protocol is realized through two complementary, lightweight modules. First, \textbf{Online Correction Memory (OCM)} acts as the proposal module by exploiting the heavy-tailed (Pareto) distribution of rejection patterns. It maintains a compact memory of frequent divergence pairs to propose alternative candidates during rejections. Second, \textbf{Semantic Consistency Gating (SCG)} serves as the verification module, mitigating the risk of proposal relaxation. By enforcing a confidence-based acceptance criterion derived from the target model, SCG filters out low-confidence substitutions that may lead to inconsistencies. Together, these modules form a unified inference protocol (Algorithm~\ref{alg:csd_step}), allowing CSD to recover benign rejections on the fly without retraining.

\begin{algorithm}[t!]
\caption{Calibrated Speculative Decoding with Online Adaptation}
\label{alg:csd_step}
\begin{algorithmic}[1]
\REQUIRE Target and Draft Models $M_p, M_q$, Input Prefix $x$, Lookahead Steps $\gamma$, Temperature $T$, Memory $\mathcal{T}$ (OCM), Thresholds $\tau$ (SCG), $\lambda$ (Freq.)

\FOR{$i = 1$ to $\gamma$}
    \STATE $q_i(x) \leftarrow M_q^{(T)}(\cdot \mid x, \tilde{x}_{<i})$ 
    \STATE $\tilde{x}_i \sim q_i(x)$
\ENDFOR

\STATE $p_{1:\gamma+1}(\cdot) \leftarrow M_p^{(T)}(\cdot \mid x, \tilde{x}_{1:\gamma})$
\STATE $n \leftarrow 0$

\FOR{$i = 1$ to $\gamma$}
    \STATE $r \sim U[0, 1]$
    \STATE $\alpha \leftarrow \min(1, \frac{p_i(\tilde{x}_i)}{q_i(\tilde{x}_i)})$
    
    \IF{$r \le \alpha$}
        \STATE $n \leftarrow n + 1$
    \ELSE
        \STATE $p'_{res}(x) \leftarrow \text{norm}(\max(0, p_i(x) - q_i(x)))$\STATE $t^* \sim p'_{res}(x)$ 
        \STATE \textcolor{blue}{\COMMENT{$\triangleright$ CSD Logic: Attempt Rescue}}
        
        \STATE $\mathit{is\_freq} \leftarrow (\mathcal{T}[(\tilde{x}_i, t^*)] \ge \lambda)$
        \STATE $\mathit{is\_safe} \leftarrow (\frac{p_i^{(T=1)}(\tilde{x}_i)}{p_i^{(T=1)}(t^*)} \ge \tau)$ 

        \STATE \textsc{UpdateMemory}$(\mathcal{T}, \tilde{x}_i, t^*)$
        
        \IF{$\mathit{is\_freq} \land \mathit{is\_safe}$}
            \STATE $n \leftarrow n + 1$ \COMMENT{\textcolor{blue}{\textbf{Rescued!}}}
        \ELSE
            \STATE \textbf{return} $x \oplus [\tilde{x}_1, \dots, \tilde{x}_n, t^*]$ 
        \ENDIF
    \ENDIF
\ENDFOR

\STATE $t \sim p_{\gamma+1}(x)$
\RETURN $x \oplus [\tilde{x}_1, \dots, \tilde{x}_\gamma, t]$

\end{algorithmic}
\end{algorithm}



\subsection{Online Correction Memory}
\label{sec:ocm}
OCM instantiates the \emph{Frequency-Guided Candidate Selection} stage of CSD. It records frequently observed divergence pairs $(\tilde{x}_i, t^*)$, where the draft token $\tilde{x}_i$ differs lexically from the verified target token $t^*$, yet such divergences recur systematically during speculative decoding.
These recurring patterns serve as empirical priors for proposing alternative candidates under future rejections.
The workflow of OCM is divided into two phases.
\paragraph{Phase 1: Prior Calibration (Initialization).}
Before deployment, we perform a lightweight calibration using standard speculative decoding on an unlabeled task corpus.
This process collects decoding-level divergence statistics to initialize the memory table $\mathcal{T}$.

Importantly, this calibration does not involve parameter updates and only captures co-occurrence statistics between the draft and target models. It serves to establish a generic prior for the model pair, rather than task-specific tuning.


\paragraph{Phase 2: Online Evolution (Inference).} During inference, OCM adapts to the test-time distribution via two mechanisms. First, \textbf{Dynamic Update} increments the frequency of any verified divergence $(\tilde{x}_i, t^*)$ in $\mathcal{T}$ (Algorithm~\ref{alg:csd_step}, Line~18), enabling CSD to capture domain-specific vocabulary or stylistic shifts online. Second, \textbf{Candidate Proposal} ensures a divergence is proposed as a rescue candidate only if it satisfies a frequency threshold $\mathcal{T}[(\tilde{x}_i, t^*)] \ge \lambda$ (Line~16). This filtering suppresses stochastic mismatches and focuses on systematic patterns, consistent with observations in Section~\ref{sec:empirical_analysis}.

\subsection{Semantic Consistency Gating}
\label{sec:scg}
While OCM leverages historical priors to propose alternative candidates, these priors are inherently context-agnostic. However, token validity in language generation is strongly conditioned on the local context, and a frequent historical substitution may still be inappropriate in a given instance.

To bridge the gap between static historical priors and dynamic decoding contexts, SCG instantiates the \textit{Probability-Guarded Acceptance} principle by implementing a context-aware verification mechanism that validates each correction candidate conditioned on the target model's instantaneous confidence

\begin{table*}[t]
\setlength{\abovecaptionskip}{3pt}
\setlength{\belowcaptionskip}{3pt}
\centering
\small
\caption{\textbf{Main results} across Llama-3 (70B/1B) and Qwen-2.5 (72B/7B) families. We report \textbf{Acc} (Task accuracy), \textbf{Tp} (Throughput in tokens/s), and \textbf{Spd} (Speedup relative to vanilla decoding). \textbf{Avg.} represents the mean speedup across all benchmarks. Best results are highlighted in \textbf{bold}.}
\label{tab:main_results}

\small                         
\setlength{\tabcolsep}{0pt}    

\begin{tabular*}{\textwidth}{@{\extracolsep{\fill}}lccccccccccccc}
\toprule

& \multicolumn{3}{c}{\textbf{GSM8K}} 
& \multicolumn{3}{c}{\textbf{MATH500}} 
& \multicolumn{3}{c}{\textbf{HumanEval}} 
& \multicolumn{3}{c}{\textbf{CNN/DM}} 
& \textbf{Avg.} \\

& \multicolumn{3}{c}{\scriptsize (8-shot, Pass@1)} 
& \multicolumn{3}{c}{\scriptsize (4-shot, Pass@1)} 
& \multicolumn{3}{c}{\scriptsize (0-shot, Pass@1)} 
& \multicolumn{3}{c}{\scriptsize (0-shot, ROUGE-L)}
& \\

\cmidrule(lr){2-4} \cmidrule(lr){5-7} \cmidrule(lr){8-10} \cmidrule(lr){11-13} \cmidrule(l){14-14}

\textbf{Method} 
& \textbf{Acc} & \textbf{Tp} & \textbf{Spd} 
& \textbf{Acc} & \textbf{Tp} & \textbf{Spd} 
& \textbf{Acc} & \textbf{Tp} & \textbf{Spd} 
& \textbf{Acc} & \textbf{Tp} & \textbf{Spd} 
& \textbf{Spd} \\
\midrule

\multicolumn{14}{l}{\textit{\textbf{Llama-3 Series}}} \\
\rule{0pt}{2.5ex}%
Vanilla Decoding 
& 92.6 & 13.4 & 1.00$\times$ & 46.0 & 15.2 & 1.00$\times$ & 76.8 & 15.9 & 1.00$\times$ & 20.3 & 15.3 & 1.00$\times$ & 1.00$\times$ \\

SpecDecode        
& 92.3 & 23.3 & 1.74$\times$ & 45.4 & 28.8 & 1.89$\times$ & 76.8 & 30.1 & 1.90$\times$ & 20.3 & 22.8 & 1.49$\times$ & 1.75$\times$ \\

Lossy SD ($\tau=0.6$)        
& 92.3 & 24.2 & 1.80$\times$ & \textbf{49.4} & 30.6 & 2.00$\times$ & 78.0 & 32.3 & 2.04$\times$ & 20.3 & 23.6 & 1.55$\times$ & 1.85$\times$ \\

SWIFT        
& 92.6 & 12.9 & 0.96$\times$ & 46.4 & 16.5 & 1.09$\times$ & 76.8 & 17.5 & 1.10$\times$ & 20.2 & 17.5 & 1.14$\times$ & 1.07$\times$ \\

Lookahead        
& 92.6 & 11.1 & 0.83$\times$ & 46.6 & 12.5 & 0.82$\times$ & 76.8 & 14.4 & 0.91$\times$ & 20.3 & 12.5 & 0.82$\times$ & 0.84$\times$ \\

\textbf{CSD (Ours)} 
& \textbf{92.6} & \textbf{24.5} & \textbf{1.83$\times$} & 48.0 & \textbf{35.5} & \textbf{2.33$\times$} & \textbf{79.3} & \textbf{37.0} & \textbf{2.33$\times$} & \textbf{20.3} & \textbf{24.4} & \textbf{1.59$\times$} & \textbf{2.02$\times$} \\

\midrule 

\multicolumn{14}{l}{\textit{\textbf{Qwen-2.5 Series}}} \\
\rule{0pt}{2.5ex}%
Vanilla Decoding 
& 91.7 & 13.9 & 1.00$\times$ & 82.2 & 15.4 & 1.00$\times$ & 88.4 & 14.9 & 1.00$\times$ & 19.9 & 15.2 & 1.00$\times$ & 1.00$\times$ \\

SpecDecode        
& 91.5 & 22.2 & 1.59$\times$ & 81.4 & 29.6 & 1.92$\times$ & \textbf{89.0} & 26.9 & 1.81$\times$ & 19.9 & 20.2 & 1.33$\times$ & 1.66$\times$ \\

\textbf{CSD (Ours)} 
& \textbf{92.2} & \textbf{25.2} & \textbf{1.81$\times$} & \textbf{83.2} & \textbf{32.8} & \textbf{2.12$\times$} & 88.4 & \textbf{29.0} & \textbf{1.95$\times$} & \textbf{19.9} & \textbf{23.7} & \textbf{1.57$\times$} & \textbf{1.86$\times$} \\

\bottomrule
\end{tabular*}
\end{table*}

\paragraph{Semantic Admissibility Check.}
SCG assesses whether a proposed token $\tilde{x}_i$ lies within a reasonable confidence margin relative to the target model's preferred token $t^*$.
Rather than computing explicit probability ratios, we perform this check directly in the logit space for efficiency:
\begin{equation}
    \label{eq:scg_check}
    \log \left( \frac{p_i^{(T=1)}(\tilde{x}_i)}{p_i^{(T=1)}(t^*)} \right) = \underbrace{z_i(\tilde{x}_i) - z_i(t^*)}_{\text{Zero-overhead}} \ge \log \tau
\end{equation}
where $z_i(\cdot)$ denotes the raw logits produced by the target model.
This formulation avoids redundant Softmax computations, is invariant to sampling temperature, and incurs negligible computational overhead.

\paragraph{Gating Logic.}
We adopt a relatively lenient threshold $\tau$, acknowledging that valid continuations are not restricted to the single top-ranked token. By allowing candidates with non-negligible confidence under the target model, SCG accommodates benign lexical variations while effectively rejecting low-confidence deviations that may cause contextual inconsistency or hallucination.

\section{Experiments}
\label{sec:experiments}

\begin{table*}[t]
\vspace{-0.3cm}
\setlength{\abovecaptionskip}{3pt}
\setlength{\belowcaptionskip}{3pt}
\centering
\scriptsize 
\caption{\textbf{Comparison with Semantic Verification Methods.} \textbf{AR} denotes the Acceptance Rate. Other metrics (\textbf{Acc}, \textbf{Tp}, \textbf{Spd}) follow the same definitions as in Table \ref{tab:main_results}. Best results are highlighted in \textbf{bold}.}
\label{tab:semantic_verification_results}

\setlength{\tabcolsep}{0pt}

\begin{tabular*}{\textwidth}{@{\extracolsep{\fill}}l cccc cccc cccc cccc cccc}
\toprule

& \multicolumn{4}{c}{\textbf{GSM8K}} 
& \multicolumn{4}{c}{\textbf{MATH500}} 
& \multicolumn{4}{c}{\textbf{HumanEval}} 
& \multicolumn{4}{c}{\textbf{CNN/DM}} 
& \multicolumn{4}{c}{\textbf{Avg.}} \\

\cmidrule(lr){2-5} \cmidrule(lr){6-9} \cmidrule(lr){10-13} \cmidrule(lr){14-17} \cmidrule(l){18-21}

\textbf{Method} 
& \textbf{Acc} & \textbf{Tp} & \textbf{Spd} & \textbf{AR} 
& \textbf{Acc} & \textbf{Tp} & \textbf{Spd} & \textbf{AR} 
& \textbf{Acc} & \textbf{Tp} & \textbf{Spd} & \textbf{AR} 
& \textbf{Acc} & \textbf{Tp} & \textbf{Spd} & \textbf{AR} 
& \textbf{Acc} & \textbf{Tp} & \textbf{Spd} & \textbf{AR} \\
\midrule

Vanilla Decoding & 92.6 & 13.4 & 1.00$\times$ & 0.0\% & 46.0 & 15.2 & 1.00$\times$ & 0.0\% & 76.8 & 15.9 & 1.00$\times$ & 0.0\% & \textbf{20.3} & 15.3 & 1.00$\times$ & 0.0\% & 58.9 & 14.9 & 1.00$\times$ & 0.0\% \\

SpecDecode & 92.3 & 22.0 & 1.64$\times$ & 46.2\% & 45.4 & 28.2 & 1.85$\times$ & 42.9\% & 76.8 & 25.6 & 1.61$\times$ & 39.9\% & 20.3 & 17.2 & 1.13$\times$ & 23.2\% & 58.7 & 23.2 & 1.56$\times$ & 38.0\% \\

Fly & 91.8 & 23.7 & 1.77$\times$ & 53.1\% & 46.8 & 32.0 & 2.10$\times$ & 51.4\% & 76.8 & 31.3 & 1.97$\times$ & 48.2\% & 20.3 & 17.5 & 1.14$\times$ & 26.7\% & 58.9 & 26.1 & 1.75$\times$ & 44.8\% \\

Reflect Verification & 92.0 & 22.6 & 1.69$\times$ & \textbf{62.0}\% & \textbf{49.6} & 29.6 & 1.94$\times$ & 58.2\% & \textbf{78.0} & 30.6 & 1.93$\times$ & \textbf{51.7}\% & 20.2 & \textbf{22.7} & \textbf{1.49$\times$} & \textbf{38.7\%} & \textbf{59.9} & 26.4 & 1.76$\times$ & \textbf{52.6\%} \\

\textbf{CSD} & \textbf{92.5} & \textbf{24.6} & \textbf{1.84$\times$} & 58.3\% & 49.4 & \textbf{35.2} & \textbf{2.31$\times$} & \textbf{59.3\%} & 76.8 & \textbf{33.4} & \textbf{2.11$\times$} & 49.8\% & 20.2 & 20.2 & 1.32$\times$ & 33.3\% & 59.7 & \textbf{28.4} & \textbf{1.89$\times$} & 50.2\% \\

\bottomrule
\end{tabular*}
\end{table*}

\begin{table*}[t]
\setlength{\abovecaptionskip}{3pt}
\setlength{\belowcaptionskip}{3pt}
\centering
\small
\caption{\textbf{Ablation study} of CSD components on MATH500 and HumanEval. \textbf{AR} denotes the Acceptance Rate. Note that individual modules (\textit{w/} OCM or \textit{w/} SCG) incur accuracy degradation compared to the baseline, while our integrated CSD framework achieves optimal balance.}
\label{tab:ablation}

\small 
\setlength{\tabcolsep}{0pt} 

\begin{tabular*}{\textwidth}{@{\extracolsep{\fill}}lcccccccc}
\toprule
& \multicolumn{4}{c}{\textbf{MATH500}} & \multicolumn{4}{c}{\textbf{HumanEval}} \\
\cmidrule(lr){2-5} \cmidrule(lr){6-9}
\textbf{Variant} & \textbf{Acc} & \textbf{AR} & \textbf{Tp} & \textbf{Spd} & \textbf{Acc} & \textbf{AR} & \textbf{Tp} & \textbf{Spd} \\
\midrule

SpecDecode (Baseline) & 45.4 & 63.6\% & 28.8 & 1.00$\times$ & 76.8 & 59.7\% & 30.1 & 1.00$\times$ \\

SD \textbf{w/} OCM & 37.8 & 83.1\% & 36.6 & 1.27$\times$ & 70.7 & 71.6\% & 37.2 & 1.24$\times$ \\

SD \textbf{w/} SCG & 43.6 & 88.7\% & 38.2 & 1.33$\times$ & 70.7 & 88.6\% & 44.4 & 1.48$\times$ \\

\textbf{CSD (Ours)} & \textbf{48.0} & \textbf{79.6\%} & \textbf{35.5} & \textbf{1.23$\times$} & \textbf{79.3} & \textbf{67.9\%} & \textbf{37.0} & \textbf{1.23$\times$} \\

\bottomrule
\end{tabular*}
\end{table*}

\subsection{Experimental Setup}
\label{sec:setup}

\paragraph{Models \& Benchmarks.}
To evaluate the versatility of CSD, we employ two representative model configurations:
(1) Llama series \citep{dubey2024llama}, pairing \texttt{Llama-3-70B-Instruct} (Target) with \texttt{Llama-3.2-1B-Instruct} (Draft); 
and (2) Qwen-2.5 series \citep{qwen2025qwen25technicalreport}, pairing \texttt{Qwen-2.5-72B-Instruct} (Target) with \texttt{Qwen-2.5-7B-Instruct} (Draft).
Standardized evaluations are conducted using the \texttt{lm-evaluation-harness} \citep{eval-harness} on four datasets spanning diverse capabilities: 
GSM8K \citep{cobbe2021training} and Minerva-Math500 \citep{hendrycks2021measuring,lewkowycz2022solving} for mathematical reasoning, 
HumanEval \citep{chen2021evaluating} for code generation, 
and CNN/DailyMail \citep{hermann2015teaching} for summarization.

\paragraph{Baselines.}
We evaluate CSD against a suite of training-free decoding strategies across four categories:
First, as standard benchmarks for strictly lossless generation, we employ Vanilla Decoding and Standard Speculative Decoding (SpecDecode) \citep{leviathan2023,chen2023accelerating}. Second, to investigate the limits of memory-free probability relaxation, we evaluate Static Lossy Speculative Decoding (Static Lossy SD), which solely employs confidence gating (Eq.~\ref{eq:scg_check}). In this setting, we tune the threshold $\tau \approx 0.6$ to identify the \textit{``lossless boundary''}—the most aggressive relaxation possible without degrading downstream performance. Third, we compare against advanced acceleration methods, including Swift \citep{xia2024swift}, which implements on-the-fly self-speculation via adaptive layer-skipping and tree-based decoding, and Lookahead Decoding \citep{fu2024break}, which achieves draft-free acceleration through parallel Jacobi iteration. We further compare CSD with recent concurrent methods: \textit{Fly} \citep{li2025training}, which employs entropy-based gating and deferred consistency validation; and \textit{Reflective Verification} \citep{wang2025think}, which utilizes reflective prompting and dual-copy templates to fuse multi-level logits.

\paragraph{Implementation Details.}
We implement CSD using PyTorch and Hugging Face Transformers \citep{paszke2019pytorch, wolf2020transformers}. All experiments are conducted on a server node equipped with 8 NVIDIA H20 GPUs. For each inference session, we allocate 2 GPUs to load the 70B/72B target models. All evaluations are performed in a single-batch setting.

\paragraph{Main Evaluation Protocol:} For the primary results (Section \ref{sec:mainres}), we set the CSD lookahead steps to $\gamma=6$, frequency threshold to $\lambda=6$, and semantic gating threshold to $\tau=0.01$.

\paragraph{Semantic Baseline Protocol:} For a fair comparison with concurrent methods (Section~\ref{sec:semantic_baselines}), we unify the speculative length to $\gamma=15$, aligning with \textit{Fly}'s default setting. Since official implementations are currently unavailable, we meticulously re-implemented these frameworks: for \textit{Fly}, we set the deferred window to 6 and optimized its entropy threshold to 0.05; for \textit{Reflective Verification}, we strictly adhered to the original hyperparameter configurations. This unified protocol ensures an equitable evaluation of CSD against the latest semantic-aware paradigms.

\textbf{Calibration:} CSD is initialized with a calibration set of 2,000 samples for most tasks, extending to 8,000 for MATH500 to ensure comprehensive coverage. We sample from the training sets of GSM8K, CNN/DailyMail, and original MATH (for MATH500). For HumanEval, we strictly use the MBPP training set \citep{austin2021program} as a proxy to ensure zero-shot integrity and prevent data leakage. During this phase, we set the temperature $T=0.6$ to broaden the CSD's exposure to diverse valid token pairs. Importantly, this calibration is a one-time, offline procedure that introduces zero online overhead during inference. The process is highly efficient, taking approximately 1.5 hours per 1,000 samples on two H20 GPUs.

\textbf{Evaluation:} 
To ensure reproducibility and eliminate stochastic variance in speedup measurements, all final performance evaluations are conducted using greedy decoding ($T=0$).

\subsection{Main Results}
\label{sec:mainres}

Table~\ref{tab:main_results} compares CSD against baselines across Llama-3 and Qwen-2.5 families. Overall, CSD consistently yields the highest throughput while maintaining target model quality.

\textbf{Performance on Llama-3.} CSD outperforms standard speculative decoding on the Llama-3-70B/1B pair. It achieves an average speedup of \textbf{2.02$\times$}, surpassing SpecDecode\footnotemark{} (1.75$\times$) by a wide margin, with gains peaking at \textbf{2.33$\times$} on MATH500 and HumanEval. Additional evaluations on complex instruction-following (IFEval) and long-context (RULER) benchmarks further validate CSD's generalization (see Appendix~\ref{sec:app_ifeval_ruler}). Crucially, CSD maintains the accuracy of vanilla decoding across most standard benchmarks, while notably surpassing the baseline on HumanEval (\textbf{+2.5} points) and MATH500 (\textbf{+2.0} points). We hypothesize that the draft model enables the system to escape the local optima of greedy decoding by proposing valid alternative trajectories.\footnotetext{Standard speculative decoding is theoretically lossless; however, minor fluctuations (e.g., 92.6 vs. 92.3 on GSM8K) may arise from non-deterministic parallel reduction kernels during verification.}

\textbf{Advantage over Lossy SD.} The performance of \textit{Lossy SD} ($\tau=0.6$) confirms the prevalence of "neutral divergence"—valid tokens that differ from the greedy path. However, static gating lacks granularity. A fixed global threshold cannot perfectly distinguish acceptable deviations from errors, necessitating a conservative setting ($\tau=0.6$) to ensure safety. This rigidity limits its ability to accept lower-probability but valid tokens, resulting in only marginal speedups over standard SpecDecode (e.g., \textbf{1.85$\times$} vs. 1.75$\times$ on Avg). CSD leverages fine-grained filtering to unlock significantly higher throughput (Avg. \textbf{2.02$\times$}).

\textbf{Comparison with Advanced Methods.} 
Complex acceleration schemes often yield suboptimal results on large-scale models. Lookahead can even yield negative speedups: its Jacobi decoding reduces the number of steps but incurs excessive FLOPs, creating a compute bottleneck on 70B models. 
Similarly, SWIFT achieves only modest gains ($\sim$1.07$\times$) due to a structural trade-off between minimizing speculation overhead and maintaining acceptance rate. For example, a low layer-skip rate (e.g., 45\%) results in a computationally heavy draft model, while a higher skip rate causes the acceptance rate to collapse. In contrast, CSD improves the acceptance rate with negligible overhead, allowing these gains to translate directly into higher throughput.

\begin{figure*}[ht]
\vspace{-0.5cm}
\setlength{\abovecaptionskip}{3pt}
\setlength{\belowcaptionskip}{3pt}
  \centering
  \includegraphics[width=\textwidth]{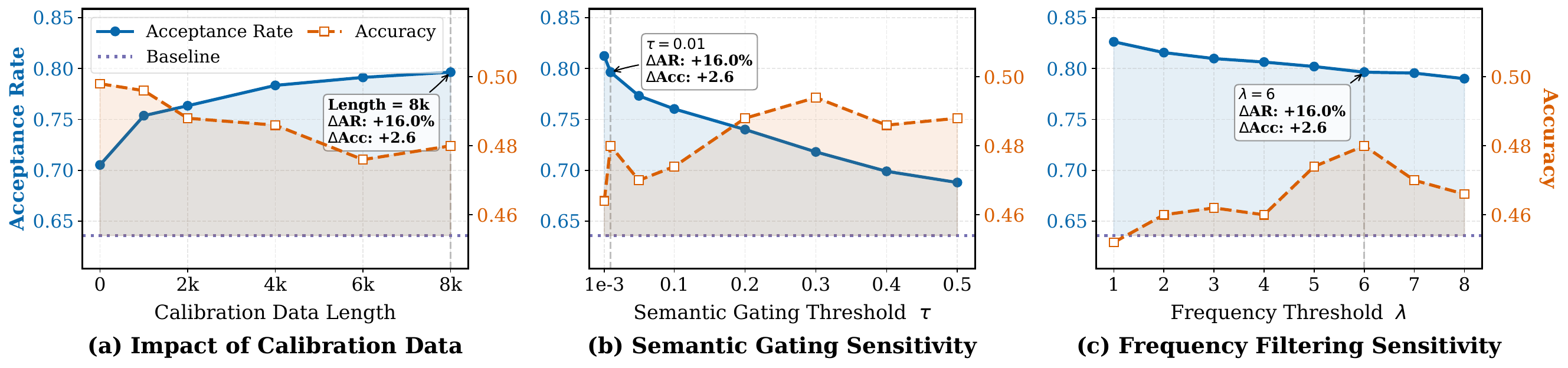} 
\caption{\textbf{Sensitivity analysis on MATH500.} (a) \textbf{Calibration}: AR plateaus beyond 2k--4k samples and peaks at 8k. (b) \textbf{Semantic Gating}: Optimal trade-off at $\tau=0.01$. (c) \textbf{Frequency}: $\lambda=6$ maximizes accuracy. Dotted lines indicate the accuracy and AR baselines of SpecDecode.}
    \label{fig:sense}
\end{figure*}

\textbf{Generalization to Qwen-2.5.} 
On Qwen-2.5 series (72B/7B), CSD also raises the average speedup from \textbf{1.66$\times$} to \textbf{1.86$\times$}. 
Notably, CSD peaked on MATH500 with a \textbf{2.12$\times$} speedup and \textbf{+1.0 point} accuracy gain. This confirms that our strategy generalizes well to different vocabularies and structures.

\textbf{Analysis of Rescued Tokens.}
To characterize the specific sources of CSD's efficiency gains, we analyzed 100 randomly sampled ``rescued'' tokens from the Llama-3-70B/1B inference traces. We categorize the false rejections mitigated by CSD into four primary types of semantically neutral divergences: \textbf{(1) Math Formatting} ($\approx$45\%), such as equivalent delimiters (e.g., \verb|$$| vs.\ \verb|\[|) or trailing punctuation (e.g., \verb|$,| vs.\ \verb|$|); \textbf{(2) Punctuation \& Spacing} ($\approx$20\%), including minor variations like commas versus conjunctions (e.g., ``\verb|,|'' vs.\ ``and'') or line breaks (e.g., \verb|\n| vs.\ \verb|\n\n|); \textbf{(3) Lexical Synonyms} ($\approx$20\%), where draft predictions are semantically identical (e.g., ``get'' vs.\ ``obtain''); and \textbf{(4) Reasoning Connectives} ($\approx$15\%), such as structural transitional words (e.g., ``So'' vs.\ ``Therefore''). Standard speculative decoding routinely penalizes these valid, mutually replaceable tokens. By safely salvaging these surface-level equivalents, CSD effectively breaks the acceptance rate ceiling of strict exact-match rules without disrupting the foundational reasoning process or downstream accuracy.

\subsection{Comparison with Concurrent Semantic Baselines}
\label{sec:semantic_baselines}

We evaluate CSD against state-of-the-art semantic baselines using a Llama-3-70B/1B pair ($\gamma=15$). Table~\ref{tab:semantic_verification_results} shows that CSD achieves higher end-to-end throughput by bypassing the structural and computational bottlenecks of prior paradigms.

\textbf{(1) Efficiency over Window-based Verification (vs. \textit{Fly}):} \textit{Fly} relies on a deferred window that requires exact-match consistency for subsequent $W$ tokens. This strategy is inherently sensitive to minor stylistic variations and often fails at sequence boundaries where future context is unavailable. In contrast, CSD performs token-independent semantic validation via statistical gating. This granularity allows CSD to recover valid drafts even at boundaries, achieving a superior average acceptance rate (50.2\% vs. 44.8\%).

\textbf{(2) Efficiency over Prompt-based Verification (vs. \textit{Reflective}):}
While \textit{Reflective Verification} achieves a high acceptance rate (52.6\%) through logit fusion, its wall-clock speedup (1.76$\times$) lags behind CSD (1.89$\times$). This gap stems from its use of verification templates that duplicate tokens and append probes, which significantly increases input length and computational latency during the target model's forward pass. CSD avoids this overhead by directly utilizing standard target logits, ensuring that improved acceptance rates translate directly into physical acceleration.

In summary, CSD shifts the semantic verification bottleneck from complex structural dependencies to a simple, zero-overhead filtering problem.


\subsection{Ablation Results}
\label{sec:abres}
To assess the contribution of each component, we evaluate two variants: (1) \textbf{SD w/ OCM}, employing only the Online Correction Memory; and (2) \textbf{SD w/ SCG}, using only Semantic Consistency Gating with $\tau=0.01$.

\textbf{Risks of Coarse-grained Filtering.} As shown in Table~\ref{tab:ablation}, while individual modules improve throughput, they trigger significant accuracy regressions. Specifically, SD w/ SCG achieves the maximum speedup ($1.48\times$ on HumanEval) by surging acceptance rate to 88.6\%, but at the cost of reducing accuracy to $70.7$. Similarly, SD w/ OCM enhances MATH500 throughput (AR 83.1\%) but fails to maintain precision, with accuracy dropping to $37.8$. These results underscore the risk of coarse-grained filtering: relaxing verification without the joint protection of semantic gating and frequency-guided proposals introduces harmful hallucinations.

\textbf{Safe Acceleration through Synergy.} In contrast, the full \textbf{CSD} framework effectively synergizes both modules. By adaptively identifying ``neutral divergence,'' CSD boosts the acceptance rate by 16.0\% (from 63.6\% to 79.6\%) on MATH500 while further boosting accuracy to \textbf{48.0}. Similarly, on HumanEval, it improves the acceptance rate to 67.9\% (vs. 59.7\% baseline) while achieving superior accuracy (\textbf{79.3}). These results demonstrate that CSD's fine-grained approach safely unlocks valid alternative trajectories, providing an optimal trade-off between inference speed and reasoning integrity.

\subsection{Sensitivity Analysis}
\label{sec:analysis}
We evaluate the impact of CSD's core hyperparameters on the MATH500 benchmark. 
As shown in Fig.~\ref{fig:sense}(a), the acceptance rate (AR) increases with the amount of calibration data. While the performance gain starts to plateau after 2k--4k samples, it reaches the maximum at 8k samples. Importantly, even without any initial calibration data, CSD achieves a 7\% improvement in AR over standard speculative decoding, thanks to its online dynamic update mechanism. This suggests that CSD can effectively capture the divergence patterns with minimal offline overhead. Furthermore, Fig.~\ref{fig:sense}(b) and (c) demonstrate the robustness of our filtering mechanism. CSD achieves an optimal balance between acceleration and reasoning integrity at $\tau=0.01$ and $\lambda=6$. Notably, across a wide range of thresholds, CSD consistently maintains higher accuracy and AR than the standard SpecDecode baseline (dotted lines), confirming its reliability and ease of deployment.

\begin{table}[t]
\centering
\small
\caption{Generalization to Universal Calibration. We compare CSD calibrated on task-specific data (\textbf{Spec.}) against CSD using a \textbf{Univ.} (Universal) calibration set. \textbf{Avg.} denotes the macro-average across all tasks.}
\label{tab:universal_generalization}
\setlength{\tabcolsep}{1.1pt} 
\begin{tabular*}{\columnwidth}{@{\extracolsep{\fill}}lcccccc}
\toprule
& \multicolumn{2}{c}{\textbf{Acc}} & \multicolumn{2}{c}{\textbf{AR}} & \multicolumn{2}{c}{\textbf{Speedup}} \\
\cmidrule(lr){2-3} \cmidrule(lr){4-5} \cmidrule(lr){6-7}
\textbf{Task} & \text{Spec.} & \text{Univ.} & \text{Spec.} & \text{Univ.} & \text{Spec.} & \text{Univ.} \\
\midrule
HumanEval & 79.2 & 78.0 & 67.9\% & 71.1\% & 2.33$\times$ & 2.45$\times$ \\
MATH500   & 48.0 & 50.0 & 79.6\% & 74.3\% & 2.32$\times$ & 2.23$\times$ \\
GSM8K     & 92.6 & 92.8 & 76.4\% & 74.5\% & 1.83$\times$ & 1.80$\times$ \\
CNN/DM    & 20.3 & 20.3 & 55.3\% & 54.4\% & 1.59$\times$ & 1.59$\times$ \\
\midrule
\textbf{Avg.} & \textbf{60.0} & \textbf{60.3} & \textbf{69.8\%} & \textbf{68.6\%} & \textbf{2.02$\times$} & \textbf{2.02$\times$} \\
\bottomrule
\end{tabular*}
\end{table}

\subsection{Generalization of Calibration}
To evaluate cross-domain robustness, we calibrate CSD using a Universal Calibration Set of 6,000 sequences from six diverse RedPajama \citep{weber2024redpajama} sub-domains. This set is strictly disjoint from all evaluation benchmarks.
As shown in Table~\ref{tab:universal_generalization}, downstream accuracy is rigorously preserved across all tasks. While transitioning to this general corpus causes marginal acceptance rate fluctuations, the average speedup remains exceptionally stable at 2.02$\times$, matching task-specific calibration. Notably, performance on HumanEval improves, likely due to the diverse syntactic patterns in RedPajama compared to the narrow MBPP domain. These results confirm that CSD avoids overfitting and effectively captures ubiquitous linguistic redundancies, enabling a single offline calibration to accelerate downstream applications without re-tuning.

\subsection{Analysis of Runtime Overhead}
\label{sec:appendix_overhead}

\begin{figure}[t!]
\centering
\includegraphics[width=0.48\textwidth]{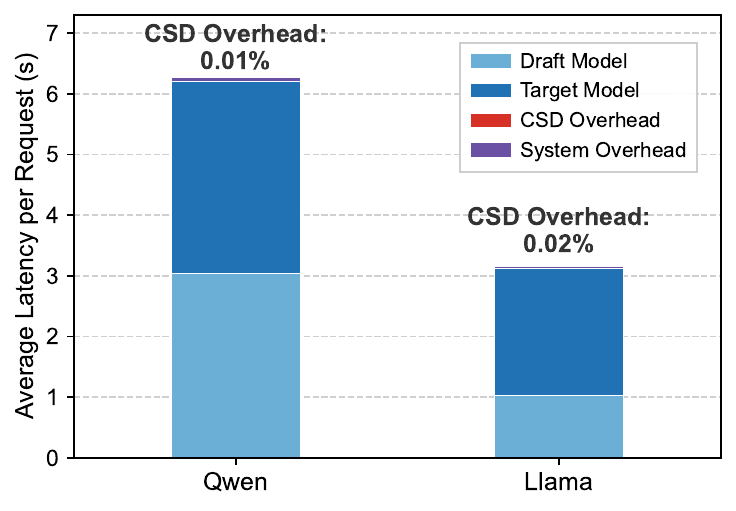} 
    \caption{\textbf{Inference Latency Breakdown.} A detailed latency analysis for Qwen-72B/7B and Llama-70B/1B. The results highlight that the CSD overhead is negligible relative to the total inference time.}
    \label{fig:brk}
\end{figure}

To evaluate computational efficiency, we profile end-to-end latency on 50 GSM8K requests (4-shot CoT). Using high-precision timing, we isolate the steady-state latency for each component. As shown in \Cref{fig:brk}, CSD's algorithmic overhead accounts for a negligible \textbf{0.01\%--0.02\%} of the total time. This confirms that our method introduces virtually no extra cost, ensuring that any increase in acceptance rate results in actual speedup.

\section{Conclusion}
\label{sec:conclusion}
We present CSD, a lightweight, training-free framework that recovers valid tokens from false rejections in SD. It combines OCM, which captures recurring divergence patterns, with SCG, which verifies candidates using model confidence rather than exact matching. 
Experiments show that CSD improves acceptance rates while maintaining accuracy and incurring negligible overhead.

\section*{Limitations}

\begin{itemize}
    \item \textbf{Departure from Distributional Exactness:} Unlike standard speculative decoding which uses rejection sampling to guarantee that output distributions are identical to the target model, CSD employs a heuristic acceptance criterion. This relaxation prioritizes inference speed over strict statistical parity. While empirical results are strong, the performance of this heuristic may not consistently generalize to unseen tasks or novel domains.
    
    \item \textbf{Dependency on Draft Quality:} The effectiveness of CSD is highly sensitive to the draft model's output quality. In instances where the draft model generates unreliable or hallucinatory sequences, the candidates fail to pass the semantic gating threshold. Such divergence leads to a significant drop in the acceptance rate, limiting the practical speedup to near auto-regressive levels.
    
    \item \textbf{High-Concurrency Scalability:} Our evaluation primarily focuses on latency reduction in small-batch settings. The mechanism for coordinating Online Correction Memory (OCM)—specifically the synchronization of real-time frequency updates across multiple concurrent requests---remains to be fully explored in high-throughput environments. The impact of potential contention in these critical sections on system-level performance warrants further investigation.Furthermore, CSD currently lacks integration with industrial-grade inference engines (e.g., vLLM), which is necessary to realize its full potential in large-scale, production-ready deployments.
\end{itemize}

\section*{Acknowledgements}
This work was partially supported by the National Key Research and Development Program of China (2024YFE0204300), National Natural Science Foundation of China (Grant No.62402311), Natural Science Foundation of Shanghai (Grant No.24ZR1433700), Key Research and Development Program of Shanghai (25LN3201200), and Alibaba Innovative Research Program. Fangxin Liu and Li Jiang are corresponding authors.

\bibliography{custom}

\appendix

\section{Extended Evaluation on Instruction Following and Long-Context Tasks}
\label{sec:app_ifeval_ruler}

To further demonstrate the versatility of CSD across highly complex scenarios, we conducted additional evaluations on IFEval \citep{zhou2023instructionfollowing} (measuring strict instruction-following capabilities) and RULER \citep{hsieh2024ruler} (evaluating long-context understanding). These experiments were conducted on the Llama-3-70B/1B model pair, with all other experimental settings kept strictly consistent with our main evaluation protocol.

As shown in Table \ref{tab:ifeval_ruler}, CSD excels on both benchmarks. On IFEval, CSD achieves a 1.77$\times$ speedup while perfectly preserving vanilla decoding accuracy (76.8). On the RULER benchmark, massive prefill overheads typically bottleneck end-to-end latency for all decoding methods. However, CSD's superior acceptance rate effectively mitigates this bottleneck, yielding a 1.37$\times$ speedup (compared to a marginal 1.02$\times$ for Standard SD) while notably boosting accuracy to 95.9.

\begin{table}[h]
\centering
\small
\caption{Performance comparison on IFEval (Instruction Following) and RULER (Long-Context). \textbf{AR} denotes Acceptance Rate, and \textbf{Spd} denotes the end-to-end speedup relative to Vanilla.}
\label{tab:ifeval_ruler}
\resizebox{\columnwidth}{!}{
\begin{tabular}{lcccccc}
\toprule
& \multicolumn{3}{c}{\textbf{IFEval}} & \multicolumn{3}{c}{\textbf{RULER}} \\
\cmidrule(lr){2-4} \cmidrule(lr){5-7}
\textbf{Method} & \textbf{Acc} & \textbf{AR} & \textbf{Spd} & \textbf{Acc} & \textbf{AR} & \textbf{Spd} \\
\midrule
Vanilla       & 76.8 & - & 1.00$\times$ & 88.1 & - & 1.00$\times$ \\
Standard SD   & \textbf{77.4} & 39.7\% & 1.65$\times$ & 88.0 & 73.6\% & 1.02$\times$ \\
CSD (Ours)    & 76.8 & \textbf{42.3\%} & \textbf{1.77$\times$} & \textbf{95.9} & \textbf{83.2\%} & \textbf{1.37$\times$} \\
\bottomrule
\end{tabular}%
} 
\end{table}

\end{document}